\documentclass[preprint,12pt,authoryear]{elsarticle}

\usepackage{amssymb}
\usepackage{amsmath}
\usepackage{booktabs}
\usepackage[table]{xcolor}
\usepackage{array}
\usepackage[ruled,vlined]{algorithm2e}
\usepackage{tabularx}
\usepackage{makecell}
\usepackage{graphicx}
\usepackage{multirow}

\usepackage{lineno}

\journal{Remote Sensing of Environment}

\begin{document}

\begin{frontmatter}



\title{Global Building Area Estimation Products: How Accurate Are They?} 


\author[label1]{Saad Lahrichi} 
\author[label1]{Doa'a Allabadi} 
\author[label2]{Kyle Bradbury} 
\author[label1]{Jordan Malof} 

\affiliation[label1]{organization={University of Missouri},
            addressline={Department of Electrical Engineering and Computer Science}, 
            city={Columbia},
            postcode={65201}, 
            state={MO},
            country={USA}}

\affiliation[label2]{organization={Duke University},
            addressline={Department of Electrical and Computer Engineering}, 
            city={Durham},
            postcode={27708}, 
            state={NC},
            country={USA}}

\begin{abstract}

Geo-spatial rasters of building footprint area are useful for a variety of tasks, such as monitoring urbanization, improving energy efficiency, and tracking greenhouse gas emissions.  There are now multiple global building raster datasets, however there lacks an independent, comprehensive, and fair assessment of their accuracy. In this work, we evaluate the accuracy of four major global building products: Global Human Settlement Layer (GHSL), Microsoft's TEMPO (TEMPO), The Global Building Atlas (GBA), and Overture.  As ground truth for assessing their accuracy, we use ORBITaL-Net, a globally diverse dataset of manually labeled building footprints. To ensure fairness, we evaluate products on grids of multiple spatial resolutions, and several conventional performance metrics.  Our results indicate that either GBA or TEMPO generally achieves the highest overall accuracy, depending upon the particular evaluation criteria. We also stratify the accuracy of each product by several factors: geographic location, population density, and income groups.  The results reveal that product accuracy can sometimes vary significantly with respect to these factors.  Notably, all products are significantly less accurate in Africa and Asia.  Most products also suffer significant accuracy reduction in high-density urban areas.    

\end{abstract}




\end{frontmatter}



\section{Introduction}
\label{sec:intro}

%

Recent advances in remote sensing, machine learning, and other technologies have made it possible to construct maps of total building footprint area at relatively high spatial resolution, but with global coverage. Building area products of this kind are useful for a variety of downstream analyses, including urbanization monitoring (e.g., quantifying urban land expansion patterns) \citep{seto2011meta}, energy efficiency (e.g., estimating the influence of building density on energy use) \citep{guneralp2017global}, disaster preparedness (e.g., quantifying exposed buildings for risk assessment) \citep{yepes2023global}, and climate change monitoring (e.g., using the building area as a proxy for greenhouse gas emissions) \citep{janssens2019edgar, markakis2026estimating, lancellotti2025closing}.  A number of global building products have recently emerged with varying key properties, allowing one to choose the most appropriate product depending upon the needs of a given application. Table \ref{tab:building_products} summarizes recent global building area estimation products, along with some of their key properties, which we briefly describe. 

\begin{table}[h]
\small
\centering
\caption{Summary of global building-related products. 
Products used in this work are highlighted in gray. *While Overture Maps has monthly data releases, its building vintages are heterogeneous and not equivalent to a temporal remote-sensing product such as TEMPO.}
\label{tab:building_products}

\resizebox{\textwidth}{!}{
\begin{tabular}{p{3cm}p{3.2cm}p{1.8cm}p{1.4cm}p{1.8cm}p{1.9cm}}
\toprule
\textbf{Product} & \textbf{Type} & \textbf{CRS} & \textbf{Res.} & \textbf{Temp. Coverage} & \textbf{Temp. Res.} \\
\midrule

\rowcolor{gray!15}
GHS-BUILT-S & Raster \newline (Building fraction) & ESRI:54009 \newline EPSG:4326 & 100 m \newline 3 arcsec & 1975--2030  & 5 years \\

\rowcolor{gray!15}
Global Building Atlas (GBA) & Vector (Footprints) & EPSG:3857 & -- & 2019 & Static \\

\rowcolor{gray!15}
Overture Maps & Vector (Footprints) & EPSG:4326 & -- & Versioned releases & Monthly* \\

\rowcolor{gray!15}
TEMPO & Raster (Building density) & EPSG:3857 & $\sim$77 m & 2018--2025 & Quarterly \\

3D-GloBFP & Vector (Footprints) & EPSG:4326 & -- & 2014-2021 & Static \\

\bottomrule
\end{tabular}
}
\end{table}

\begin{figure}
    \centering
    \includegraphics[width=0.9\linewidth]{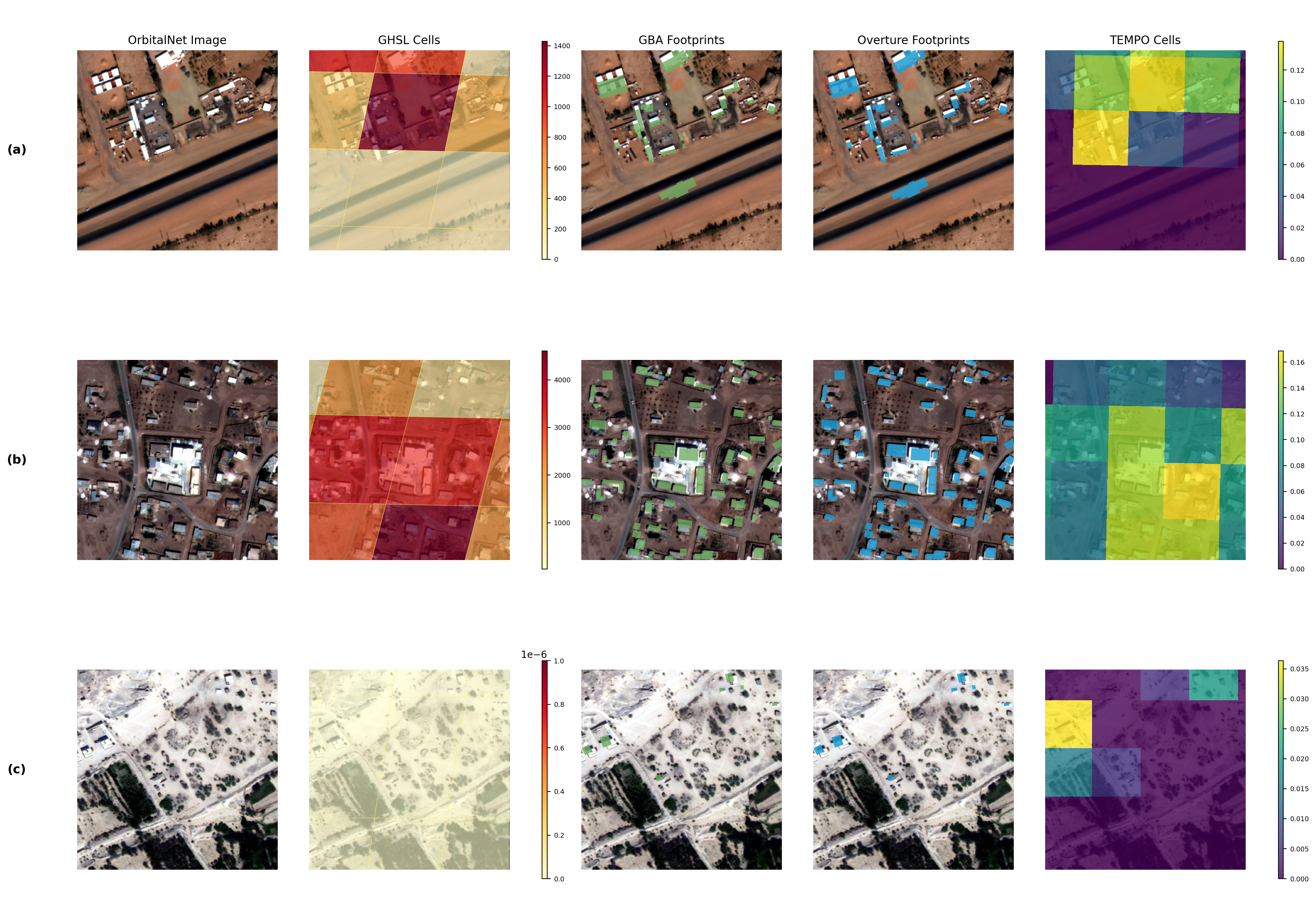}
    \caption{Three examples (a-c) of satellite imagery (first column) and corresponding building area estimates from each of the products (remaining columns) evaluated in our study.}
    \label{fig:dataset_examples}
\end{figure}

Existing building data is typically provided in one of two "Types": geospatial rasters or polygons.  Raster-based data comprises a geospatial grid wherein the total building area or density in each grid cell is reported, while vector-based products provide geospatial polygons for each building.  A key property of raster-based approaches is their spatial resolution (i.e., the area represented by each pixel), while vector-based approaches have the advantage that they can be aggregated into any gridded resolution. Figure \ref{fig:dataset_examples} shows examples of data from each of the products used in this study, over several geographic locations. Each product also uses a coordinate reference system (CRS) to represent coordinates on Earth. Each CRS has unique properties. For example, EPSG:3857, Web Mercator, represents coordinates in projected meters, and does not have equal-area grid cells. On the other hand, ESRI:54009, World Mollweide, represents coordinates in meters and has equal-area cells. Existing global building products all rely to some degree on dated satellite imagery, which defines their temporal coverage. While most products use single snapshots to release one static version, some products periodically release new versions, either using interpolation/extrapolation methods or by relying on updated imagery. 

One key property of building products that remains difficult to assess, but is crucial to support downstream applications, is their accuracy.  The absolute accuracy of a building product is useful for determining its viability for a given downstream application (e.g., due to tolerance for error), as well as estimating error in downstream products and decisions that rely on building area estimates. Accuracy is also useful for choosing the most accurate product in scenarios where multiple building products are viable.  The main contribution of this work is to estimate and compare the accuracy of recent global building area products: specifically,those in the shaded rows of Table \ref{tab:building_products}.  Our work builds upon many existing assessments of building product accuracy, by addressing some of their key limitations.


\subsection{Limitations of Existing Building Product Comparisons}

Table \ref{tab:assessment_comparison} summarizes key properties of recent research in which the accuracy of global building area products was assessed.  We next describe each of the properties (columns) in Table \ref{tab:assessment_comparison} and how our comparison often improves upon existing work.   

\textit{Ground Truth Data Accuracy.} Perhaps the most important limitation of existing comparisons is the accuracy of their ground truth.  The current gold standard ground truth source for existing product comparisons is obtained via manual annotation of individual building footprints in optical overhead (e.g., satellite) imagery. The annotation accuracy relies heavily upon the spatial resolution of the imagery because it is easier to determine the presence of a building and its precise extent in higher-resolution imagery.  The ground truth data used in this work (ORBITaL-Net \citep{swan2025orbital} - see Section \ref{sub:orbital-net}) comprises manual (human) annotation of building polygons over very high-resolution optical satellite imagery (approximately 0.47 m, in median \citep{swan2025orbital}). Furthermore, each annotation was reviewed by a second annotator and then a third expert to further improve accuracy.  Most other ground truth data utilized substantially lower-resolution satellite imagery (4 m resolution imagery, or lower), as reported in Table \ref{tab:assessment_comparison}(Source Imagery Resolution).   


\textit{Ground Truth Representativeness.} A major limitation of existing comparisons is the limited representativeness of the ground truth data.  Specifically, the ground truth data is often limited in its total size, and geographic diversity. This is important because building area products utilize diverse methodologies, and their error rates may vary with respect to geography (e.g., due to building sizes, shapes, and appearance in satellite imagery), or other sources of variability. Therefore it is useful to have ground truth data that encompasses a large and diverse geographic area. Table \ref{tab:assessment_comparison} (Total Area) reports the total area encompassed by the ground truth in recent comparisons. Table \ref{tab:assessment_comparison}(Urban and Rural?) reports whether the ground truth utilized in comparisons is diverse and includes both urban and rural areas, or is only focused on either. The work here utilizes ground truth over a substantially larger and more diverse geographic region compared to prior work, in principle yielding more robust estimates of accuracy.  

\textit{Assessment on a Single Reference Grid.} To facilitate comparisons of existing building area products, it is necessary to reproject all products onto a shared grid (e.g., CRS and resolution), termed here as a \textit{reference grid}. As we elaborate in Sec. \ref{sec:experimental_design}, and our results corroborate, comparison on a single reference grid can systematically favor some building area products over others.  To facilitate a richer and fairer comparison, it is useful to compare products on multiple reference grids. Table \ref{tab:assessment_comparison}(multi-grid) indicates which recent existing accuracy assessments utilized more than one reference grid in their assessments. 

\textit{Inclusion of Recent Existing Products.} One limitation of existing work relates to the number of relevant products included in their comparisons. Table \ref{tab:assessment_comparison}(No. of Products) reports the number of building area products from Table \ref{tab:building_products} that were included in their accuracy assessment.  The focus of this work is the accuracy of global building area products, and to our knowledge, Table \ref{tab:building_products} comprises all such products at present. 

\begin{table*}[t]
\centering
\small
\resizebox{1\textwidth}{!}{
\begin{tabular}{lccccc}
\toprule
& \textbf{Ground Truth} & \multicolumn{2}{c}{\textbf{Coverage of Ground Truth}} & \textbf{Multi-grid} & \textbf{\# Products} \\
\cmidrule(lr){3-4}
\textbf{Assessment} & \textbf{\makecell{Source Imagery\\Resolution}} & \textbf{\makecell{Total\\Area}} & \textbf{\makecell{Urban\\+ Rural?}} & \textbf{Comparisons?} & \textbf{Compared} \\
\midrule
This work  & 0.45 m & $\sim$6,480 km$^2$ & Yes & Yes & 4 \\
WSF2015  & 0.15--1.5 m & $\sim$90 km$^2$ & Yes & No & 1 \\
GHS-BUILT-S & \makecell{10 m;\\NR VHR}& \makecell{NR;\\$\sim 313 km^2$}& Yes & Yes & 0\\
TEMPO & 4 m & $\sim 991 km^2$ & No & No & 3 \\
GBA  & NR & NR & No & No & 1 \\
\bottomrule
\end{tabular}
}
\caption{Comparison of our assessment with prior studies. Ground-truth accuracy is summarized by whether the assessment uses manual building annotations and by the spatial resolution of the imagery underlying the reference data. Ground-truth coverage is summarized by total evaluated area and whether the reference data includes both urban and rural regions. NR = not reported.}
\label{tab:assessment_comparison}
\end{table*}








\subsection{Summary of Contributions of this Work}  
In this work, we perform a comprehensive and independent accuracy evaluation of recent global building area estimation products that addresses most limitations of existing assessments.  Underpinning our evaluation is access to ground truth of unprecedented quality, quantity, and geographic diversity. For ground truth building area data we leverage ORBITaL-Net \citep{swan2025orbital}, a recently released global dataset of quality-controlled building footprint masks.  In contrast to most prior work, and to provide a fairer and richer comparison, we evaluate the products on multiple raster grids, and using several different performance metrics. We also stratify our evaluation by several factors: geographic location, population density, and income groups. Our results reveal that either GBA or TEMPO achieve the highest overall accuracy depending upon the precise evaluation criteria. Furthermore, each product's accuracy can sometimes vary significantly by region and socioeconomic context, with all products performing worse in Africa, Asia, and (most) in high-density urban areas.
The remainder of this paper is organized as follows. Section \ref{sub:orbital-net} presents the ground-truth, Section \ref{sec:methods} presents the building products for comparison; Section \ref{sec:experimental_design} details our comparison methodology;Section \ref{sec:results} explains our results; and Section \ref{sec:conclusion} concludes.

\section{Ground Truth Source: ORBITaL-Net}
\label{sub:orbital-net}

To estimate the accuracy of each product, we first establish a set of reliable ground truth labels, against which we compare each product's building footprint estimates. We use the recently released ORBITaL-Net product \citep{swan2025orbital} as our ground truth data, which provides manually labeled high-resolution Maxar imagery.  ORBITaL-Net is, to our knowledge, the largest global, open, manually-labelled building dataset. It consists of 128,000 500×500 px chips and 1.49 M hand-labeled buildings derived from 0.45 m resolution Maxar satellite images collected between 2010 and 2020, with 2017 as the median year.  Since it was developed to enable the training of generalizable machine learning models to predict buildings globally, it includes imagery from various geographies across the globe, including both rural and urban areas. To create this dataset, the authors performed their own pre-processing (georegistration, orthorectification, and pan-sharpening) on raw Maxar satellite imagery, then used an automatic algorithm \citep{swan2022iterative} to select image scenes for manual labeling, which capture the diversity of global building appearances and imaging conditions. The final dataset includes 40-45\% of positive samples (i.e., image chips containing buildings) and 55-60\% negative samples (i.e., image chips with no buildings present). 

In addition to being the largest dataset of its kind, it also employed a well-defined validation process to ensure label quality. A team of 30 analysts labeled all visible building portions, including facades, using Maxar imagery with varied properties (brightness, contrast, band combinations, etc.) to disambiguate structures. Public imagery served as additional reference to learn regional specifics. As part of the quality assurance, the analysts followed a three-tier process: self-review, peer-review with additional training using recurring mistakes, and final expert validation. While onerous, this manual pipeline using high-resolution satellite imagery produces highly accurate labels, suitable for use as ground truth in evaluating other building products. 

\section{Building Area Estimation Products}
\label{sec:methods}

Here, we present details about each building product being compared. These products are summarized in Table \ref{tab:building_products}. While we focus in this work on building area, some of these products include height data as well, and could be used, in principle, for building volume estimation. We exclude 3D-GloBFP \citep{che20243d} from this study because its main contribution is 3D height estimation rather than a new 2D footprint product (which is more relevant for our work). Moreover, its 2D geometries suffer from two main limitations: (1) temporal inconsistency: since it relies on Microsoft Building Footprints, which are derived from a rolling mosaic of Bing satellite imagery, the actual acquisition dates of the building footprints can range anywhere from 2014 to 2021. (2) redundancy with other datasets: since we are only interested in 2D building area evaluation, this dataset's footprint layer is redundant with sources already incorporated into GBA and is less comprehensive, with reported regional gaps inherited from the source Microsoft data. 


\subsection{GHSL}
\label{sub:ghsl}
The Global Human Settlement Layer (GHSL), part of the European Commission's Copernicus Program, provides global data products characterizing human presence on Earth. In this work, we use GHS-BUILT-S, which provides a global gridded estimation of built-up surfaces. Unlike other datasets, instead of classifying pixels as built/not built, GHSL estimates a sub-pixel built-up surface fraction (fBU) at 10 m resolution.

For the 2018 reference epoch, a Symbolic Machine Learning (SML) model is trained on a composite of existing built-up products (GHSL, Facebook, Microsoft, OSM), and applied to Sentinel-2 image features to produce 10 m fBU estimates, which are aggregated to a 100 m grid and converted to built-up surface area per cell. For historical and future epochs, built-up surface growth is modeled using temporal extrapolation and spatial–temporal allocation using static factors (terrain, slope, elevation) and growth trends derived from observed epochs (1975, 1990, 2000, 2014 from Landsat and 2018 from Sentinel-2). For the 2020 epoch we use in our study, this results in positive growth relative to 2018, while preserving observed settlements.  

Validation against independent building footprint datasets (WSF, ESRI) that are rasterized and aggregated to match GHSL grids reports an MAE of 6\% of the grid cell area, i.e., 600 $m^2$ at 100 m resolution. The authors acknowledge remaining uncertainty in sparse rural settlements, very small and isolated buildings, and the lack of a globally uniform ground truth.

\subsection{GBA}
\label{sub:gba}
The Global Building Atlas (GBA) \citep{zhu2025globalbuildingatlas} provides building polygons (GBA.Polygon), heights (GBA.Height), and a combined product (GBA.LoD1) globally derived from 3 m resolution PlanetScope imagery. Each of the 2.75 billion building polygons is represented as a vector footprint, while the height information is represented as a raster. While not used in this work, we note that the GBA.Height product can be used in future work that integrates height data. 

To curate the data, the authors acquired PlanetScope Surface Reflectance imagery from 2019, supplementing with 2018 imagery when the former was unavailable in cloud-free format. Using OSM labels and a dataset of 42 Chinese cities \citep{cao2021deep}, they trained a UPerNet model \citep{xiao2018unified}, with ConvNeXt-Tiny \citep{liu2022convnet} as a backbone to predict binary building masks from the PlanetScope images. They then used a second network to refine the generated masks and separate adjacent buildings, applied GDAL contour tracing to convert the masks to polygons, and used the WorldCover dataset \citep{zanaga_2021_5571936} to remove false positives.

Finally, a quality-based polygon fusion strategy was used to produce the LoD1 product using the model-generated GBA.Polygon and existing open building polygons: OSM \citep{openstreetmap}, Google Open Buildings \citep{sirko2021continental}, Microsoft Building Footprints \citep{microsoft_global_ml_building_footprints}, and CLSM \citep{shi2024last}. The final product uses the highest quality dataset as a primary source (typically OSM or Google Open Buildings in the Global South) and merges it with the dataset with the highest combined recall and additional building area. The model-generated polygons serve as a globally complete dataset and are used in regions where the existing datasets have missing data.

\subsection{TEMPO}
\label{sub:microsoft}
TEMPO \citep{glazer2025tempo} provides global, quarterly maps of building density and height at 77 m resolution from Q1 2018 through Q2 2025. Unlike ORBITaL-Net and GBA, and similar to GHSL, the buildings are represented as gridded rasters, instead of building footprints. TEMPO is derived from global mosaics of PlanetScope RGB imagery. The authors train their EfficientNet-B6 U-net model using existing building footprint and height datasets as weak supervision, including Google Open Buildings 2.5D, and Overture Maps. The model jointly predicts per-pixel building density and height. 

Due to the lack of a global ground truth dataset, the TEMPO predictions are validated by assessing cross-dataset alignment with Google Open Buildings 2.5D, Overture Maps, GHSL, WSF, and GBA using a set of 59 SpaceNet7 quads and temporal consistency. The authors note that TEMPO inherits spatial bias from weak supervision, with performance varying by region, and that the single-view RGB imagery used for training can miss small buildings, low-contrast, or isolated structures. In this work, we use the publicly released 2023 Q4 global epoch for our comparison.

\subsection{Overture}
\label{sub:overture}

The Overture Maps Foundation \citep{overturemaps} provides a global, open-source dataset of building footprints generated through the integration of multiple existing geospatial datasets. Unlike products derived directly from satellite imagery using machine learning models, Overture aggregates building footprints from several sources, including OpenStreetMap and other open or commercial building footprint datasets. Buildings are represented as vector polygons distributed in the WGS84 coordinate reference system (EPSG:4326). The dataset is produced through a data conflation process that merges overlapping building footprints from different sources and removes duplicate geometries. Overture is released as versioned global snapshots; in this study we use the February 2026 release (2026-02-18.0).  Because the dataset aggregates existing building footprints rather than generating them directly from satellite imagery, its spatial coverage and completeness depend on the availability and quality of the underlying source datasets. As a result, coverage may vary across regions depending on the extent of mapping efforts and available building footprint data.

\section{Experimental Design}
\label{sec:experimental_design}
The goal of our experiments is to evaluate and compare the accuracy of each building area product listed in Table \ref{tab:building_products}(gray rows).  We first provide a high-level description of our experimental design, followed by a more detailed mathematical description.  The ground truth data for our accuracy evaluation is provided by the ORBITaL-Net dataset (see Fig. \ref{sub:orbital-net}) which comprises 500 by 500 pixel binary geospatial rasters, or \textit{masks}, indicating where buildings are present.  These annotated rasters are distributed pseudo-randomly around the globe.  To evaluate the accuracy of the building products, we first adopt a ground truth \textit{reference grid}, and then compute the total area of ORBITaL-Net building polygons within each cell of the reference grid. We then reproject all building products onto this reference grid, so that each product provides an estimate of building area over each cell of the reference grid.

\textbf{Utilizing Multiple Reference Grids.} However, the choice of reference grid (e.g., its CRS and spatial resolution) can bias the accuracy estimates in favor of products with higher-resolution grids, or grids that are more similar to the reference grid. This arises because reprojection of one product onto the reference grid requires some type of estimation (e.g., proportional allocation \citep{thomas2015accuracy}), which can introduce additional error. The estimation procedure is likely most error-prone when projecting a low-resolution grid onto a higher-resolution grid (e.g., GHSL onto the TEMPO grid). Our main accuracy evaluation results in Sec. \ref{sec:results} corroborate these assertions.  To ensure a fair comparison among multiple building products, we evaluate the accuracy of the building products several times, each time using the native grid of a different product as the reference grid.  This approach allows each product to be evaluated on its native grid once, revealing its accuracy under ideal conditions, as well as revealing other products' accuracy under non-ideal conditions.  In this work we compare all the products in Table \ref{tab:building_products}(gray rows), which includes two raster products: GHSL (available in two projections) and TEMPO, resulting in three distinct experiments: (1) the GHSL WGS84 3 arcsec grid, (2) the GHSL Mollweide 100 m grid, and (3) the TEMPO EPSG:3857 77 m grid.  

For each reference grid in our experiments, we estimate the accuracy of the products using the following general steps: (1) Computation of ground truth values on the reference grid, and (2) Projection of the building products onto the reference grid.  In step (2), we utilize two procedures: one for raster products, and another for vector products (see Table \ref{tab:building_products}). We describe each of these three steps in detail next, along with mathematical descriptions in Algorithm \ref{alg:ground_truth_reference_grid} and Algorithm \ref{alg:product_predictions_reference_grid}.     



\begin{algorithm}[H]
\LinesNumbered
\caption{Compute Ground Truth for a Reference Grid}
\label{alg:ground_truth_reference_grid}

\KwIn{Set of geospatial polygons representing regions where buildings have been annotated, $R_s=\{r_s\}$ in CRS $s$; polygons (i.e., pixels in ORBITaL-Net) where buildings exist, $B_s=\{b_s\}$, where we assume $b_s \subset r_s$ for some $r_s$; polygons (i.e., grid cells) in the desired reference grid  $P_t=\{p_t\}$, which is defined in CRS $t$.}

\KwOut{A table $T_{\mathrm{GT}}$ with reference grid cells $p_t^\star$ found to have complete ground truth coverage, and their corresponding ground-truth building areas $a_{\mathrm{GT}}(p_t^\star)$.}

\BlankLine
\tcp{Identify reference grid cells that have complete ground-truth coverage}
Initialize $P_t^\star \leftarrow \emptyset$\;
\ForEach{$r_s \in R_s$}{
    Reproject the annotated region $r_s$ into the product reference CRS $t$: 
    \[r_t=\mathcal{T}_{s \rightarrow t}\left(r_s\right)\]
    
    Find all reference grid cells $p_t$ sufficiently covered by annotated region $r_t$:
    \[P_t^\star \leftarrow P_t^\star \cup \left\{p_t \in P_t : \frac{\left|p_t \cap r_t\right|}{|p_t|} \ge 0.99 \right\}.\]
    }
\BlankLine
\tcp{Compute ground-truth building area on each selected reference cell}

\ForEach{$p_t^\star \in P_t^\star$}{

    Reproject the reference grid cell $p_t^\star$ into the annotation CRS $s$:
    \[p_s^\star = \mathcal{T}_{t \rightarrow s} \left(p_t^\star\right).\]

    Find all building-labeled pixels that live within the projected product grid cell:
    \[B_s^\star=\left\{b_s \in B_s : \operatorname{centroid}(b_s) \in p_s^\star  \right\}.\]

    Compute the ground-truth building area:
    \[a_{\mathrm{GT}}(p_t^\star) = \operatorname{area} \left(\bigcup_{b_s \in B_s^\star} b_s\right).\]

    Add $(p_t^\star, a_{\mathrm{GT}}(p_t^\star))$ to $T_{\mathrm{GT}}$.
}

\end{algorithm}

\subsection{Computing the Ground Truth Mask on the Reference Grid}
\label{sub:ghsl_contained}
\begin{figure}
    \centering
    \includegraphics[width=\linewidth]{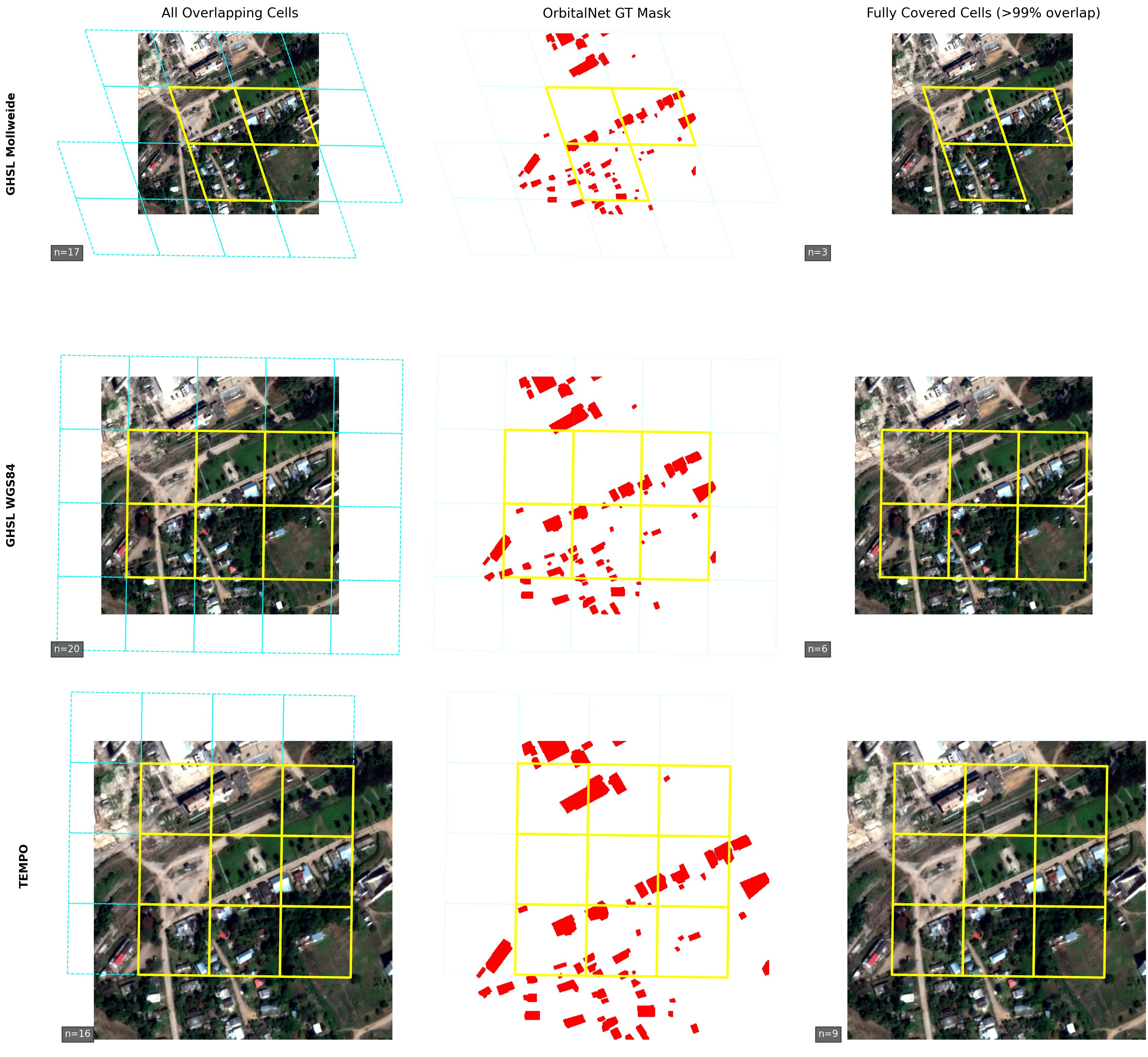}
    \caption{Left: A sample ORBITaL-Net optical image, overlaid by reference grid cells (shown in blue) that overlap with it. Center: The corresponding ORBITaL-Net building mask. Right: The same ORBITaL-Net optical image, overlaid only by the reference grid cells retained by our selection process, shown in yellow. Each row uses one of our three reference grids (GHSL Mollweide, GHSL WGS84, and TEMPO). $n$ in the bottom left of the image shows the count of overlapping/fully-contained cells. We observe that the grids do not align with one another and therefore result in different levels of coverage of the ORBITaL-Net image.}
    \label{fig:fully_covered_cells}
\end{figure}

This process is presented mathematically in Algorithm \ref{alg:ground_truth_reference_grid}, and here we provide a written description.   First, we need to identify reference grid cells that are (near-)completely covered by an ORBITaL-Net ground truth mask, so that an accurate ground truth area value can be obtained for that cell.  To do this, for each ORBITaL-Net building mask, we identify all overlapping reference grid cells and retain only those cells that are fully contained within the spatial footprint of the mask. Specifically, we reproject the ORBITaL-Net mask into the reference grid CRS. For each candidate grid cell (e.g., for GHSL WGS84, 3 arcsec x 3 arcsec) intersecting this mask, we build the corresponding grid cell polygon and reproject it back into the ORBITaL-Net CRS for building area computation. We then intersect each grid cell polygon with the ORBITaL-Net image polygon and compute an overlap ratio, defined as the intersection area divided by the cell area. We retain only cells with an overlap ratio at least 99\%, ensuring that each retained cell is fully contained within the ORBITaL-Net image footprint. This filtering step removes boundary cells that are only partially covered by ORBITaL-Net imagery, which would otherwise lead to incomplete building-area estimates. Fig. \ref{fig:fully_covered_cells} shows an example of an ORBITaL-Net image, its corresponding building mask, and the subset of reference grid cells that are fully contained within the image footprint. In this example, we observe that the number of such cells varies ($n=3$ using GHSL Mollweide, $n=6$ using GHSL WGS84, and $n=9$ using TEMPO). All subsequent cross-dataset comparisons are restricted to the set of fully contained reference grid cells.

To compute ORBITaL-Net building area within each selected grid cell, we create a boolean mask in the ORBITaL-Net grid and the image affine transform to extract the building-labeled pixels inside the cell. The ORBITaL-Net building area is then computed as  $A_{ORB} = N_{bldg} \cdot a_{px} $ where $N_{bldg}$ is the number of building-labeled pixels within the grid cell (assigned by centroid), and $a_{px}$ is the ground area of a single ORBITaL-Net pixel derived from the affine transform. This results in a single ORBITaL-Net building area estimate per grid cell, allowing direct cell-level comparison with the reference grid. Using the same set of fully contained cells, we then compute the building area from the remaining products for comparison.


\begin{algorithm}[H]
\LinesNumbered
\caption{Compute Product Predictions on a Reference Grid}
\label{alg:product_predictions_reference_grid}
\KwIn{
Selected reference grid cells $P_t^\star=\{p_t^\star\}$ from Algorithm~\ref{alg:ground_truth_reference_grid}; building products $\mathcal{D}$, each represented as either vector building polygons or raster cells with density values.
}
\KwOut{
A table $T_{\mathrm{pred}}$ containing predicted building areas $\hat{a}_D(p_t^\star)$ for each product $D \in \mathcal{D}$ and each selected reference geometry $p_t^\star$.
}
\ForEach{$p_t^\star \in P_t^\star$}{
    \ForEach{building product $D \in \mathcal{D}$}{

        Reproject $p_t^\star$ into the coordinate system of product $D$:
        \[
        p_D^\star = \mathcal{T}_{t \rightarrow D}(p_t^\star).
        \]

        \uIf{$D$ is represented by vector polygons $V_D=\{v_D\}$}{
            Compute predicted building area:
            \[\hat{a}_D(p_t^\star) = \operatorname{area}
            \left(p_D^\star \cap \bigcup_{v_D \in V_D} v_D \right).\]
            }

        \ElseIf{$D$ is represented by raster cells $S_D=\{s_i\}$}{

            Find valid source cells intersecting $p_D^\star$:
            \[S_D^\star = \left\{s_i \in S_D :s_i \cap p_D^\star \neq \emptyset
            \right\}.\]

            Aggregate source-cell densities over $p_D^\star$:
            \[\bar{d}_D(p_t^\star)=\frac{\sum_{s_i \in S_D^\star}d_D(s_i)\,|s_i \cap p_D^\star|}{\sum_{s_i \in S_D^\star}|s_i \cap p_D^\star|}.
            \]
            
            Convert aggregated density to building area:
            \[\hat{a}_D(p_t^\star)=\bar{d}_D(p_t^\star)\,|p_D^\star|.\]
        }
        Add $(p_t^\star, D, \hat{a}_D(p_t^\star))$ to $T_{\mathrm{pred}}$\;
    }
}

\end{algorithm}

\subsection{Compute Product Estimates on a Reference Grid}

This process is presented mathematically in Algorithm \ref{alg:product_predictions_reference_grid}, and here we provide a written description: one for vector products, and one for raster products.  Each of the two processes shares several steps. 

\textbf{Vector Products.} After identifying the reference grid cells where ground truth is available (i.e., Algorithm \ref{alg:ground_truth_reference_grid}), we transform the selected reference grid cells and vector building products into a common CRS, intersect the reference cells with the building footprints, and compute the building area intersecting each reference cell. This allows us to get all of the buildings (or parts of buildings) that fall within each cell. The final building area value is computed by taking the union of all intersection pieces in each cell. 

\textbf{Raster Products.} For raster products, we first express each product cell value as a building density. Products that already report density, such as TEMPO, are used directly, while products that report built-up area, such as GHSL, are converted to density by dividing the total cell building area by the cell geospatial area. We then aggregate product cell densities over each selected reference geometry using average resampling, which corresponds to area-weighted averaging of source-cell densities over each destination cell. Finally, we convert the resulting density to total building area by multiplying by the reference-cell area.

\subsection{Baseline Estimator}
We accompany the building products with a constant \textit{Baseline} estimator. Let $\mathcal{V}$ denote the set of $n$ valid reference-grid cells included in the evaluation, and let $y_i$ be the ground-truth building area in cell $i$. We define the baseline value as the mean ground-truth building area across these cells $\bar{y}=\frac{1}{n}\sum_{j \in \mathcal{V}} y_j$. The baseline assigns this same value to every reference-grid cell. Therefore, this estimator provides a simple reference against which the building products can be compared. Because it assigns the observed mean to every cell, its predicted total building area is identical to the ground-truth total, and its coefficient of determination is zero when the ground-truth values have nonzero variance. It is therefore most informative for the cell-level error metrics introduced in Table \ref{tab:evaluation_metrics}, like MAE and WMAPE.

\begin{table}[t]
\centering
\caption{Evaluation metrics used to compare building area products against ORBITaL-Net ground truth. All metrics are computed over the matched set of valid cells $S$, where $y_i$ is the ORBITaL-Net building area in cell $i$, $\hat{y}_i$ is the product estimate, $n=|S|$, and $\bar{y}$ is the mean ground-truth building area over $S$.}
\label{tab:evaluation_metrics}
\resizebox{\textwidth}{!}{
\begin{tabular}{cccp{8.5cm}}
\toprule
Acronym & Name & Equation & Explanation \\
\midrule
Total & Total building area & $\displaystyle \sum_{i \in S} \hat{y}_i$ & Measures the aggregate amount of building area predicted by a product. This provides a global sanity check on whether the product is systematically over- or under-estimating building coverage. \\

$\Delta$ & Aggregate difference  & $\displaystyle \Delta = \sum_{i \in S}(\hat{y}_i - y_i)$ & Quantifies the direction and magnitude of aggregate bias. Positive values indicate overestimation, while negative values indicate underestimation. Bias is interpretable in physical units. \\

$\Delta(\%)$ & Percentage difference & 
$\displaystyle \Delta(\%) = \frac{\sum_{i \in S}(\hat{y}_i-y_i)}{\sum_{i \in S} y_i}\times 100$ & Similar to aggregate difference, makes the bias interpretable relative to the ground-truth total. \\

MAE & Mean absolute error & $\displaystyle \frac{1}{n}\sum_{i \in S} |\hat{y}_i-y_i|$ & Measures average per-cell error in square meters. MAE is directly interpretable in the same physical units as the data and is commonly used in prior studies. \\

WMAPE & Weighted mean absolute percentage error & $\displaystyle \frac{\sum_{i \in S} |\hat{y}_i-y_i|}{\sum_{i \in S} y_i}\times 100$ & Measures total absolute error normalized by total ground-truth building area. Avoids instability from cells with zero or near-zero building area and provides a scale-normalized measure of overall error. \\

$R^2$ & Coefficient of determination & $\displaystyle 1 - \frac{\sum_{i \in S}(y_i-\hat{y}_i)^2}{\sum_{i \in S}(y_i-\bar{y})^2}$ & Measures how well a product captures spatial variation in building area across cells and is also reported in previous work. A high $R^2$ indicates that the product correctly distinguishes cells with relatively low versus high building density. Uncertainty in $R^2$ is estimated using 1000 bootstrap resamples. \\

\bottomrule
\end{tabular}
}
\end{table}

\section{Results}
\label{sec:results}

We present our results in three parts: first, each raster product is benchmarked against all others on its native reference grid. For each dataset, we report five complementary metrics that capture different aspects of dataset accuracy. The metrics are summarized in Table \ref{tab:evaluation_metrics}. Then, we visually compare qualitative examples of our products' predictions against the ground truth. Finally, we stratify the results by continent, population density, and income level. 

\subsection{Benchmark Results}
\textbf{Broad Findings.} Table \ref{tab:overall_building_comparison} summarizes the results of the comparison of GHSL (using two projections), GBA, TEMPO, and Overture using the ORBITaL-Net data ground truth under each considered reference grid. The results vary depending upon the reference grid and the performance metric considered, however, some broad trends emerge.  First, all products achieve substantially lower error than the \textit{Baseline} approach, implying that all products are superior to a random estimator.  Among the products, GHSL-based products always achieve the highest error rates across all reference grids and metrics, including the GHSL reference grids. Among the remaining three products (TEMPO, GBA, and Overture), the results are more nuanced.  GBA always performs best across all three reference grids according to the MAE and WMAPE measures. By contrast, TEMPO always performs best on all three reference grids according to the $\Delta(m^{2})$, $\Delta(\%)$, and $R^{2}$. Compared to $R^{2}$, the MAE and WMAPE metrics are less influenced by high-error predictions.  Therefore, the aforementioned results  suggest that TEMPO makes more small-magnitude errors compared to GBA and Overture, but produces fewer very large-magnitude prediction errors.  

\textbf{Building Products Exhibit Some Systemic Bias.} Interestingly, the $\Delta$ measures indicate that GBA, TEMPO, and Overture all exhibit a bias whereby they systematically (i.e., on average) underestimate the true building area, with TEMPO having the least bias.  By contrast, GHSL systematically overestimates building area, and by a larger magnitude than the other products underestimate it.  This result corroborates other literature that highlights GHSL's tendency to overestimate building area \citep{liu2020accuracy, uhl2023spatially, leyk2018assessing, glazer2025tempo}.   

\textbf{Reprojection of Raster Products Introduces Additional Error.} As noted in Sec. \ref{sec:experimental_design}, we observe that reprojecting the raster products (e.g., TEMPO and GHSL) onto other grids tends to increase their overall error.  For example, the MAE and WMAPE of TEMPO both increase, while its $R^2$ decreases when it is evaluated on the GHSL reference grids, compared to the TEMPO grid.  The error is most egregious for MAE, likely because MAE is most sensitive to many small increases in error than the other metrics, which we hypothesize is the most likely impact of the re-projections. The negative impact of GHSL reprojection onto the TEMPO grid is generally more pronounced because the TEMPO grid is higher-resolution than the GHSL grid.  The Overture and GBA products are vector-based, and therefore have the advantage that they can be projected onto any grid with (in theory) no additional error.  It is therefore likely that GBA and Overture would exhibit much lower error on higher-resolution grids (e.g., 50m or 25m) compared to the grid products. However, one advantage of the GHSL and TEMPO products is that they are periodically updated to provide more contemporary area estimates.           

\textbf{Why Does Total Building Area Change Depending Upon the Reference Grid Used?} We note that the absolute total area changes drastically depending on the reference grid (e.g., the ORBITaL-Net ground truth shifts from $\approx$ 67 million $m^2$ on the 100 m grid to $\approx$ 141 million $m^2$ on the 77 m grid). This implies that the selection process described in Section \ref{sub:ghsl_contained} and Algorithm \ref{alg:ground_truth_reference_grid} results in different amounts of valid ground truth over which we can compare. However, the relative behavior of all products ($\Delta$ \%) remains stable, which further separates the performance of the products from the choice of reference grid used. 

\begin{table}[t]
\centering
\caption{Comparison of global building products against ORBITaL-Net truth using different reference grids}
\label{tab:overall_building_comparison}
\resizebox{\textwidth}{!}{  
\begin{tabular}{l l c c c c c c}

\toprule
Ref. Grid & Product & Total (m$^2$) & MAE & WMAPE & $\Delta$ (m$^2$) & $\Delta$ (\%) & $R^2$ \\
\midrule

\multirow{6}{*}{\shortstack[l]{GHSL\\ \\3 arcsec}}
 & ORBITaL-Net & 113{,}931{,}494 & -- & --  & -- & -- & -- \\
 & Baseline    & 113{,}931{,}494 & 426.61 & 161.09  & -- & -- & -- \\
 & GHSL        & 144{,}416{,}146 & 183.50 & 54.69  & 30{,}484{,}651 & 26.75 & 0.5804 $\pm$ 0.003 \\
 & GBA         & 99{,}192{,}674 & 85.27 & 32.19  & -12{,}856{,}344 & -11.47 & 0.8108 $\pm$ 0.003 \\
 & TEMPO       & 105{,}421{,}689 & 104.22 & 39.37  & -8{,}509{,}804 & -7.47 & 0.8220 $\pm$ 0.002 \\
 & Overture    & 90{,}702{,}680 & 101.61 & 38.39  & -23{,}228{,}814 & -20.39 & 0.7336 $\pm$ 0.003 \\

\midrule

\multirow{6}{*}{\shortstack[l]{GHSL\\ \\100 m}}
 & ORBITaL-Net & 67{,}590{,}511 & -- & --  & -- & -- & -- \\
 & Baseline    & 67{,}590{,}511 & 557.01 & 160.06  & -- & -- & -- \\
 & GHSL        & 87{,}476{,}436 & 217.86 & 48.38  & 19{,}885{,}924 & 29.42 & 0.6745 $\pm$ 0.004 \\
 & GBA         & 58{,}757{,}372 & 111.33 & 31.79  & -7{,}339{,}537 & -11.10 & 0.8194 $\pm$ 0.004 \\
 & TEMPO       & 62{,}898{,}539 & 137.36 & 39.48  & -4{,}691{,}972 & -6.94 & 0.8206 $\pm$ 0.003 \\
 & Overture    & 53{,}244{,}790 & 135.91 & 39.06  & -14{,}347{,}600 & -21.23 & 0.7253 $\pm$ 0.002\\

\midrule

\multirow{6}{*}{\shortstack[l]{TEMPO\\ \\77 m}}
 & ORBITaL-Net & 141{,}742{,}174 & -- & -- & -- & -- & -- \\
 & Baseline    & 141{,}742{,}174 & 246.44 & 164.50 & -- & -- & -- \\
 & GHSL WGS84  & 175{,}907{,}502 & 117.16 & 78.19 & 34{,}285{,}842 & 24.21 & 0.5534 $\pm$ 0.002 \\
 & GHSL Moll   & 178{,}132{,}098 & 115.11 & 76.81 & 36{,}517{,}019 & 25.79 & 0.5473 $\pm$ 0.002 \\
 & GBA         & 123{,}170{,}568 & 50.89 & 34.09 & -16{,}262{,}840 & -11.66 & 0.7951 $\pm$ 0.002 \\
 & TEMPO       & 132{,}262{,}760 & 51.44 & 34.34 & -9{,}479{,}414 & -6.69 & 0.8369 $\pm$ 0.001 \\
 & Overture    & 113{,}131{,}838 & 59.78  & 39.91 & -28{,}610{,}336 & -20.18 & 0.7214 $\pm$ 0.002\\

\bottomrule
\end{tabular}
}
\end{table}

\subsection{Qualitative Results}
\begin{figure}
    \centering
    \includegraphics[width=\linewidth]{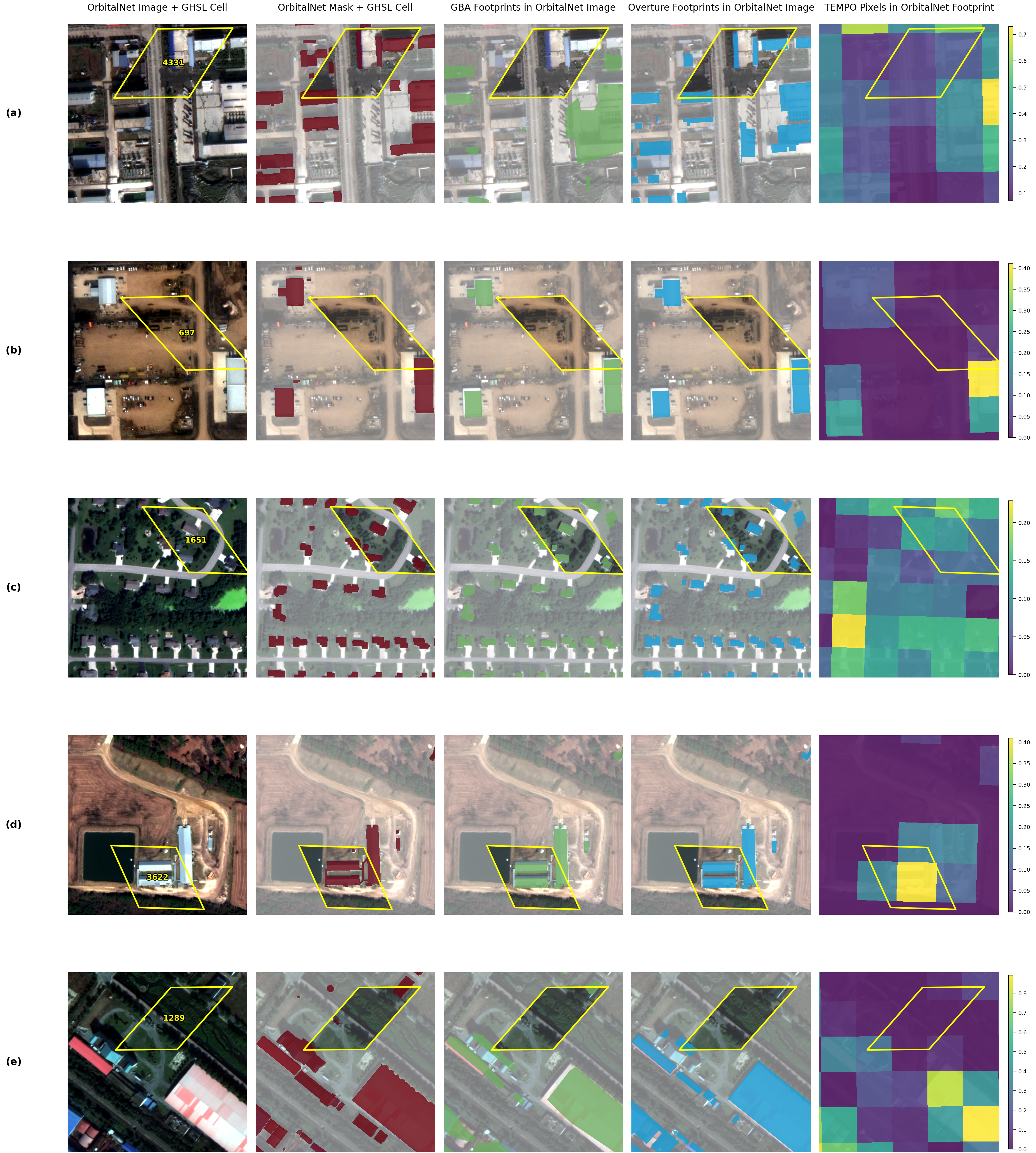}
    \caption{Qualitative samples showing the pipeline used to compare products using the GHSL Mollweide reference grid cells. The first column shows the ORBITaL-Net image overlaid with the GHSL grid cell and the GHSL value. The second column shows the corresponding binary mask. The third and fourth columns show the GBA and Overture vectors, respectively, where the area intersecting the GHSL cell is highlighted. The last column shows the TEMPO predictions and the color bar shows the values represented by each pixel color.}
    \label{fig:qualit_result_1}
\end{figure}

In Fig. \ref{fig:qualit_result_1}, we show five examples of qualitative results visualizing the ground truth ORBITaL-Net image and corresponding mask, the fully-covered GHSL cell, the GBA and Overture footprints, and the TEMPO density map for various locations. We observe that, at a high level, the datasets agree in terms of building presence and location. However, they sometimes differ in their detections. For example, in row (a), GHSL predicts 4,331 $m^2$ building area inside the cell; the ORBITaL-Net and Overture masks show high similarity both inside and outside the cell. However, GBA predicts no building area inside the cell, and its masks outside of it do not spatially match the ground truth. On the other hand, row (b) shows almost complete agreement between all datasets, on an example where buildings are sparsely distributed. The agreement is in spatial location, shape, and size of the buildings. Although the TEMPO density mask is not easily comparable visually, this example shows that the pixel covering the built area has much higher intensity (colored yellow) than the surrounding ones (colored purple), which represent non-built areas. Rows (c) and (d) also show examples of high agreement between the datasets, not only in densely built residential areas, but also in less urban areas. Finally, row (e) shows an instance where the ground truth mask includes one small building and two portions of buildings. For this sample, GBA correctly predicts the two built portions but misses the small building. Overture only predicts one of the building portions, and TEMPO shows slightly positive pixel values around the correct areas.      

\subsection{Stratifying Building Product Accuracy by Key Factors}
While aggregate metrics summarize global performance, they do not inform any regional biases or specific failure cases.  We hypothesize that building products' performance depends on the types of buildings and the surrounding context. In this section, we test this hypothesis by selecting various factors that could influence the performance of algorithms (e.g., by affecting appearance of buildings or background), and stratify our results by each of these factors. These include continents, population density, and income classification. Our earlier results indicated that GHSL consistently performed worst among products, and therefore we present a stratified analysis solely using the TEMPO reference grid here.

Breaking down our ground truth by continent reveals that TEMPO and GBA achieve the highest $R^2$ values, with TEMPO achieving the highest value in all continents but South America, where GBA achieves $R^2= 0.92$ versus 0.89 for TEMPO. Overture performs well in the Americas and Oceania, but lags behind in Africa and Asia. GHSL achieves a decent $R^2 = 0.73$ in South America, but is weaker in Asia, North America, and Oceania, with $0.42 < R^2 < 0.59$. It also fails to predict Africa well, with an $R^2 = -0.37$, showing a strong example of regional failure that's not apparent in aggregate metrics. This is also observed when comparing $R^2$ magnitudes. TEMPO, GBA, and Overture all achieve $0.81 < R^2 < 0.92$ in the Americas and Oceania, but that drops to $0.57 < R^2 < 0.77$ for Africa and Asia, highlighting differences in the regional accuracies of the products. Interestingly, we notice that aside from GBA in South America, the three non-GHSL products underpredict in all continents, with negative bias $\Delta (\%)$ except for Africa, where they all overpredict to varying degrees: GBA and Overture $< 2\%$, TEMPO $\approx 14\%$, while GHSL suffers from severe overestimation in Africa (+131\%) and moderate overestimation in North America (+41\%) and Oceania (+20\%). 

\begin{table}[h]
\centering
\caption{Performance stratified by continent}
\label{tab:continent}
\resizebox{\textwidth}{!}{
\begin{tabular}{lrrrrrrrr}
\toprule
& \multicolumn{2}{c}{TEMPO} & \multicolumn{2}{c}{GHSL} & \multicolumn{2}{c}{GBA} & \multicolumn{2}{c}{Overture}\\
\cmidrule(lr){2-3} \cmidrule(lr){4-5} \cmidrule(lr){6-7} \cmidrule(lr){8-9}
Continent & $R^2$ & $\Delta$ (\%) & $R^2$ & $\Delta$ (\%) & $R^2$ & $\Delta$ (\%) & $R^2$ & $\Delta$ (\%) \\
\midrule
Africa        & 0.76 & 13.80  & -0.37 & 131.46 & 0.72 & 1.84   & 0.62 & 0.41 \\
Asia          & 0.77 & -10.86 & 0.59  & 9.63   & 0.70 & -20.64 & 0.57 & -36.60 \\
North America & 0.90 & -3.12  & 0.48  & 40.51  & 0.89 & -3.87  & 0.88 & -4.89 \\
Oceania       & 0.86 & -2.71  & 0.42  & 19.84  & 0.81 & -4.81  & 0.83 & -5.04 \\
South America & 0.89 & -7.55  & 0.73  & 8.19   & 0.92 & 2.17   & 0.85 & -12.74 \\
\bottomrule
\end{tabular}
}
\end{table}
Next, we examine each product's performance as a function of population density. We use the WorldPop dataset \citep{tatem2017worldpop} to sample population values at each cell, and adapt the European Commission's degree of urbanization framework \citep{european2021applying} to define population density bins. Specifically, we stratify cells into very low-density rural cells ($<50$ people/km$^2$), low-density rural cells ($50$--$300$ people/km$^2$), moderate-density settlement cells ($300$--$1{,}500$ people/km$^2$), and high-density urban cells ($\geq1{,}500$ people/km$^2$). We note that these are not official Degree of Urbanization classes, since we do not enforce the framework's contiguity or minimum population size requirements. Rather, they provide a standardized split of density values to evaluate building product robustness to population density.

Across all population-density strata, TEMPO achieves consistently high $R^2$ ($\approx 0.8$ in very low, low, and moderate density areas, and $\approx 0.7$ in high-density urban areas. It also underpredicts by less than $3\%$ in all but high density urban areas, where the non-GHSL products' performance degrades. GBA's $R^2$ shrinks from $0.75$ in low and moderate density areas to $0.64$, and we observe a similar pattern with Overture ($0.67 \rightarrow 0.46$). On the other hand, GHSL's best performance is achieved in high density urban areas ($R^2 = 0.41$), while it does worse in very low and low density areas  ($R^2 \approx 0.35$) and is worst in moderate densities  ($R^2 = 0.28$). We also observe that TEMPO, GBA, and Overture generally underestimate across strata, especially in high density urban areas and that GBA and Overture underestimate more than TEMPO in very low, low, and moderate density areas (between -9\% and -20\%). These results underscore that errors are not only regional, but also related to population density. 

\begin{table}[h]
\centering
\caption{Performance stratified by population-density strata adapted from the Degree of Urbanisation framework}

\label{tab:pop_quantiles}
\resizebox{\textwidth}{!}{
\begin{tabular}{lrrrrrrrr}
\toprule
& \multicolumn{2}{c}{TEMPO} & \multicolumn{2}{c}{GHSL} & \multicolumn{2}{c}{GBA} & \multicolumn{2}{c}{Overture}\\
\cmidrule(lr){2-3} \cmidrule(lr){4-5} \cmidrule(lr){6-7} \cmidrule(lr){8-9}
Stratum & $R^2$ & $\Delta$ (\%) & $R^2$ & $\Delta$ (\%) & $R^2$ & $\Delta$ (\%) & $R^2$ & $\Delta$ (\%) \\
\midrule
Very Low Density Rural & 0.79 & -2.89 & 0.36 & 37.32 & 0.68 & -11.65 & 0.66	& -18.36  \\
Low Density Rural      & 0.81 & -2.56 & 0.35 & 31.71 & 0.75 &-11.69 & 0.67 &-19.73 \\
Moderate Density       & 0.80 &  -2.97 & 0.28 & 32.38 & 0.75 & -8.65 & 0.67	& -15.13 \\
High Density Urban     & 0.69 & -13.02 & 0.41 & 14.67 & 0.64 & -14.03 & 0.46	& -24.97  \\
\bottomrule
\end{tabular}
}
\end{table}
We also use the World Bank's income classification \citep{worldbank_country_lending} to measure the performance of each product based on income group. The results show that non-GHSL products perform best in high-income regions (TEMPO: 0.87, GBA: 0.83, and Overture: 0.79), with TEMPO and GBA showing generally stable performance across income groups while Overture's performance degrades in the upper middle and low income groups. GHSL shows the most severe instability, with a near-zero $R^2$ in lower-middle-income regions and 0.64 in upper middle income ones. While GHSL overestimates across income groups, it does so slightly in upper middle income regions (+2.78\%) but severely in lower middle income regions (+106\%). The lower middle income region was the only one where other products also overestimated building areas by 16-19\%, while they underestimated in the remaining regions, with TEMPO showing the least underestimation, followed by GBA, and Overture.

\begin{table}[h]
\centering
\caption{Performance stratified by World Bank income group}
\label{tab:income_group}
\begin{tabular}{lrrrrrrrr}
\toprule
& \multicolumn{2}{c}{TEMPO} & \multicolumn{2}{c}{GHSL} & \multicolumn{2}{c}{GBA} & \multicolumn{2}{c}{Overture} \\
\cmidrule(lr){2-3} \cmidrule(lr){4-5} \cmidrule(lr){6-7} \cmidrule(lr){8-9}
Income group & $R^2$ & $\Delta$ (\%) & $R^2$ & $\Delta$ (\%) & $R^2$ & $\Delta$ (\%) & $R^2$ & $\Delta$ (\%) \\
\midrule
High         & 0.87 & -4.09  & 0.49 & 37.89  & 0.83 & -8.12  & 0.79 & -12.90 \\
Upper middle & 0.79 & -11.92 & 0.64 & 2.78   & 0.73 & -19.84 & 0.62 & -34.82 \\
Lower middle & 0.77 & 19.43  & 0.02 & 105.81 & 0.74 & 16.35  & 0.68 & 18.06 \\
Low          & 0.79 & -11.00 & 0.37 & 31.96  & 0.74 & -16.13 & 0.62 & -23.53 \\
\bottomrule
\end{tabular}
\end{table}

\subsection{Limitations}
\label{sub:limitations}
Some limitations of our comparison include the temporal gaps between ground truth ORBITaL-Net data (data collected between 2010 and 2020, with 2017 as the median year) and the other datasets. In theory, this would have a more severe effect on the datasets that are outside of that range and/or further from the median. For example, the TEMPO dataset is from 2023, while GBA is from 2019. As such, there exists a systematic bias in favor of GBA estimates, as they were extracted on more temporally proximate imagery than TEMPO. Future work could find measures to correct for this systematic bias. 

\begin{figure}
    \centering
    \includegraphics[width=\linewidth]{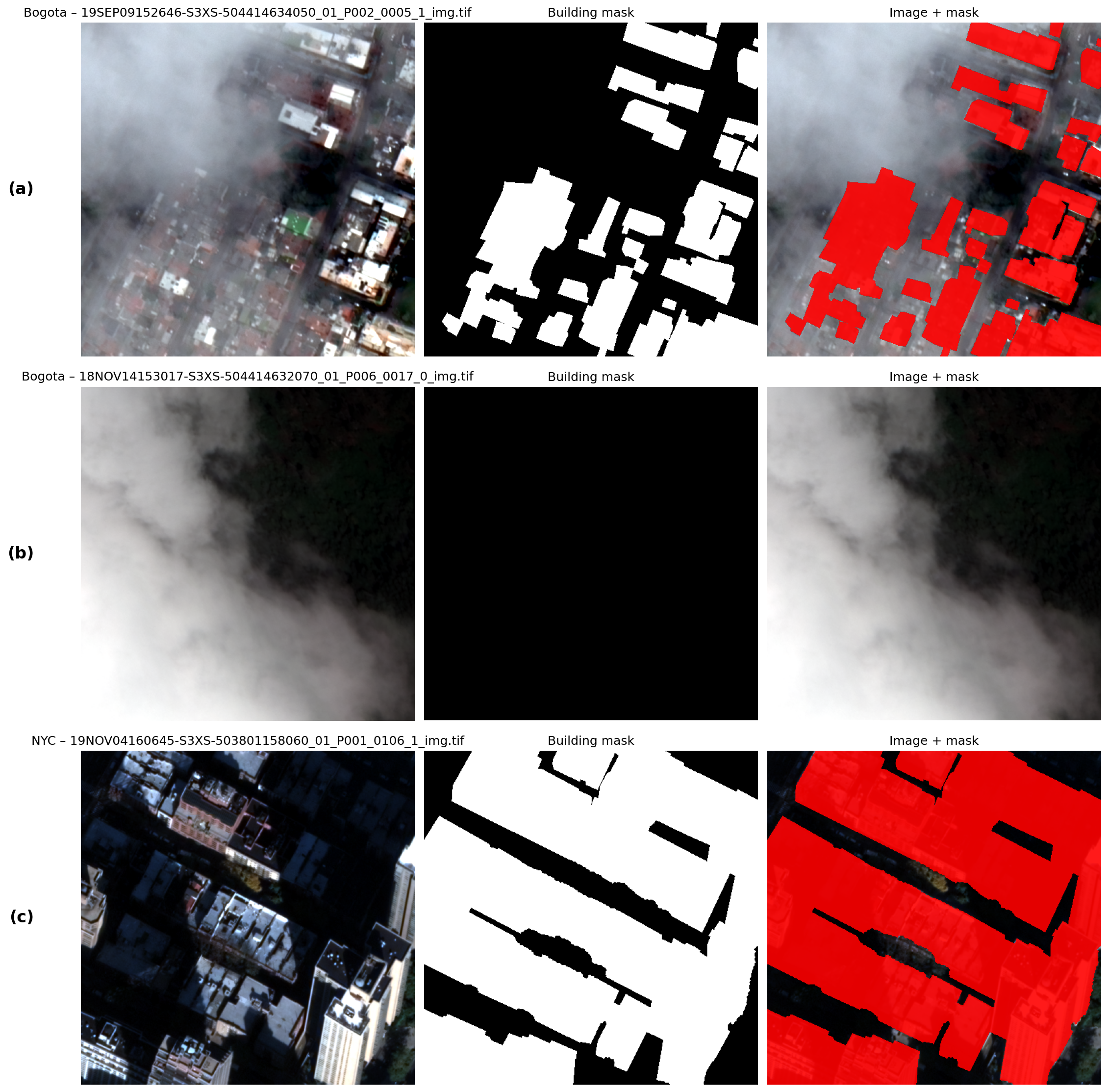}
    \caption{Examples of: (a) partially occluded imagery due to clouds, (b) fully occluded imagery due to clouds, and (c) off-nadir, partially occluded imagery due to shadows. The first column shows the ORBITaL-Net imagery, the second column shows the binary mask, and the third column overlays the mask on the imagery. }
    \label{fig:poor_examples}
\end{figure}

Another limitation is related to the definition of buildings in the data we use as ground truth. ORBITaL-Net labeled all visible parts of the building as such. As a result, there may be discrepancies between the area of building extracted from ORBITaL-Net labels and the true building area.  Such discrepancies can also arise because ORBITaL-Net was originally created to train machine learning models, rather than estimate building area, creating limitations in the labeling protocols. For example, when the imagery is off-nadir, building masks include building facaces, rather than just the rooftop. The dataset also includes images with heavy occlusion wherein buildings (if present) are not visible and therefore cannot be annotated.  In these instances the absence of building masks does not guarantee no buildings are present, potentially leading to underestimates in the ground truth. Fig. \ref{fig:poor_examples} presents examples of these labeling limitations.      

Finally, the ORBITaL-Net dataset does not include samples in Europe. As such, we are not able to compare products in that region. While including data from Europe would further strengthen our findings, we hypothesize that it may not significantly change our conclusions, given the results in North America and high income groups, where building characteristics are somewhat similar.

\section{Conclusion}
\label{sec:conclusion}
In this work, we present, to our knowledge, the first global, independent comparison of heterogeneous building area products (rasters, vectors) using manually labeled ground truth data. Our results do not identify a single product that is uniformly most accurate under every metric. GBA achieves the lowest MAE and WMAPE on all three reference grids. TEMPO achieves the smallest aggregate bias and highest $R^2$ across the same grids. Overture generally performs below these two products, while GHSL exhibits the largest errors and confirms previous results that show its high propensity for overestimation. To evaluate potential regional biases, we stratified the results using various key factors. This further confirmed that TEMPO is the most stable product across stratifications, while GHSL shows the largest bias, especially in low-population and low-income regions. GBA is generally competitive with TEMPO on aggregate, but shows some global coverage issues. By establishing a rigorous, global benchmark against manually labeled ground truth data, this work fills an important gap in the building area estimation literature and provides not only a framework for fair comparison of heterogeneous building products, but also essential context to guide dataset selection and future development.

\bibliographystyle{elsarticle-harv} 
\bibliography{references.bib}

\end{document}